\documentclass[table]{ai2style/ai2}

\usepackage{amssymb}
\usepackage{multirow}
\usepackage{bigdelim}
\usepackage{todonotes}
\usepackage{longtable}
\usepackage{tabularray}
\usepackage{wrapfig}
\usepackage[most]{tcolorbox} 
\usepackage{url}
\usepackage{xspace}
\usepackage{svg}
\usepackage[absolute]{textpos}
\usepackage{fdsymbol}
\usepackage{csvsimple}
\usepackage{booktabs}
\usepackage{amsfonts}
\usepackage{nicefrac}
\usepackage{microtype}
\usepackage[table]{xcolor}
\usepackage{amsmath}
\usepackage{csquotes}
\usepackage{siunitx}
\usepackage{graphicx}
\usepackage{arydshln}
\usepackage{enumitem}
\usepackage{soul}
\usepackage{adjustbox}
\usepackage{pifont}
\usepackage{caption}
\usepackage{makecell}
\usepackage{bold-extra}
\usepackage{booktabs,rotating,xcolor}
\usepackage{setspace}
\usepackage{quoting}
\usepackage{float}
\usepackage{nicematrix}
\usepackage{hyperref}
\usepackage{listings}
\usepackage{tikz}
\usepackage{textcomp}
\usepackage{subcaption}

\definecolor{linkcolor}{RGB}{0, 0, 128}
\hypersetup{
     colorlinks   = true,
     citecolor    = linkcolor,
     linkcolor    = linkcolor,
     urlcolor     = linkcolor,
}

\setlist[itemize]{leftmargin=*,itemsep=0em,parsep=0.3em,topsep=0.3em}

\definecolor{maroon}{HTML}{F26035}
\definecolor{yellow}{HTML}{FDBC42}
\definecolor{lavender}{HTML}{734f96}
\definecolor{darkergrey}{HTML}{444444}
\definecolor{midgrey}{HTML}{e6eded}
\definecolor{ai2pink}{HTML}{f0529c}
\definecolor{ai2midpink}{HTML}{fad3e5}
\definecolor{ai2lightpink}{HTML}{fbecf3}
\definecolor{ai2midwhite}{HTML}{f2e5d9}
\definecolor{ai2offwhite}{HTML}{fbf4ee}
\definecolor{ai2green}{HTML}{0fcb8c}
\definecolor{ai2lightgreen}{HTML}{e7f9f3}
\definecolor{ai2darkgreen}{HTML}{105257}
\definecolor{ai2purple}{HTML}{B932EB}
\definecolor{ai2lightpurple}{HTML}{f7e8fc}
\definecolor{neutralEight}{HTML}{343434}
\definecolor{neutralFive}{HTML}{838383}
\definecolor{neutralThree}{HTML}{bebebe}
\definecolor{neutralOne}{HTML}{dedede}
\definecolor{lightgrey}{HTML}{fafcfc}
\definecolor{darkred}{RGB}{156, 39, 33}
\definecolor{darkblue}{RGB}{31, 90, 153}
\definecolor{forestgreen}{rgb}{0.13, 0.55, 0.13}
\definecolor{rum_color}{HTML}{7E3DA7}
\definecolor{pi_color}{HTML}{EAB711}
\definecolor{darkgreen}{RGB}{0,100,0}

\definecolor{backcolour}{rgb}{0.95,0.95,0.92}
\lstdefinestyle{promptstyle}{
    backgroundcolor=\color{backcolour},
    basicstyle=\ttfamily\footnotesize,
    breaklines=true,
    captionpos=b,
    frame=single,
    showspaces=false,
    showstringspaces=false,
    tabsize=2,
    breakindent=0pt
}
\newcommand{\model}{\textsc{MolmoWeb}\xspace}
\newcommand{\dataset}{\textsc{MolmoWebMix}\xspace}
\newcommand{\basemodel}{\textsc{Molmo2}\xspace}
\newcommand{\webvoyagerfull}{\textsc{WebVoyager}\xspace}
\newcommand{\webvoyager}{\textsc{WebV}\xspace}
\newcommand{\omtw}{\textsc{OM2W}\xspace}
\newcommand{\omtwfull}{\textsc{Online-Mind2Web}\xspace}
\newcommand{\deepshop}{\textsc{DeepShop}\xspace}
\newcommand{\webtailfull}{\textsc{WebTailBench}\xspace}
\newcommand{\best}[1]{\textbf{#1}}
\newcommand{\sbest}[1]{\underline{#1}}

\newcommand{\eg}{e.g.\xspace}

\newif\ifshowcomments
\showcommentstrue

\ifshowcomments
\newcommand{\tg}[1]{\textcolor{red}{\textbf{TG:} #1}}
\newcommand{\yy}[1]{\textcolor{blue}{\bf\small [#1 --Yue]}}
\newcommand{\zm}[1]{\textcolor{green}{ZM: #1}}
\newcommand{\ps}[1]{\textcolor{purple}{Peter: #1}}
\newcommand{\pw}[1]{\textcolor{red}{Piper: #1}}
\newcommand{\jrc}[1]{\textcolor{orange}{JR: #1}}
\newcommand{\ranjay}[1]{\textcolor{green}{$[$#1$]^R_K$}}
\else
\newcommand{\tg}[1]{}
\newcommand{\yy}[1]{}
\newcommand{\zm}[1]{}
\newcommand{\ps}[1]{}
\newcommand{\pw}[1]{}
\newcommand{\jrc}[1]{}
\newcommand{\ranjay}[1]{}
\fi

\newcolumntype{L}[1]{>{\raggedright\let\newline\\\arraybackslash\hspace{0pt}}m{#1}}
\newcolumntype{C}[1]{>{\centering\let\newline\\\arraybackslash\hspace{0pt}}m{#1}}
\newcolumntype{R}[1]{>{\raggedleft\let\newline\\\arraybackslash\hspace{0pt}}m{#1}}
\newcolumntype{P}[1]{>{\centering\let\newline\\\arraybackslash\columncolor{ai2lightpink}}m{#1}}

\title{%
  \mbox{\includegraphics[width=0.65cm]{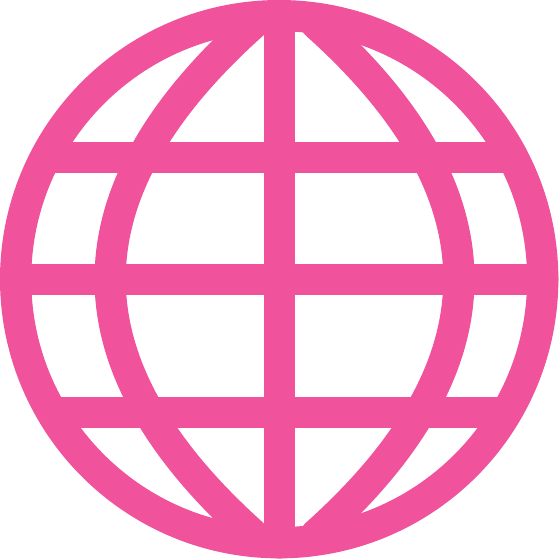}} 
   \model: Open Visual Web Agent and Open Data for the Open Web%
}

\newcommand{\core}{\textsuperscript{\textcolor{ai2pink}{\ding{170}}}}
\newcommand{\coremark}{\textcolor{ai2pink}{\ding{170}}}

\authorOne[1*]{Tanmay Gupta\core$^\dagger$}
\authorOne[1*]{Piper Wolters\core}
\authorOne[1,2*]{Zixian Ma\core}
\authorOne[1*]{Peter Sushko\core}
\authorTwo[1,2]{Rock Yuren Pang\core}
\authorTwo[1]{Diego Llanes\core}
\authorTwo[1]{Yue Yang\core}
\authorTwo[1]{Taira Anderson}
\authorTwo[1]{Boyuan Zheng}
\authorTwo[1,2,3]{Zhongzheng Ren}
\authorTwo[1]{Harsh Trivedi}
\authorTwo[1]{Taylor Blanton}
\authorTwo[1]{Caleb Ouellette}
\authorTwo[1]{Winson Han}
\authorThree[1,2]{Ali Farhadi}
\authorThree[1,2]{Ranjay Krishna\core}
\affiliation[1]{Allen Institute for AI}
\affiliation[2]{University of Washington}
\affiliation[3]{UNC-Chapel Hill}
\contribution[]{
* denotes equal technical contribution, \dagger\ indicates project lead, \coremark\ marks core contributors.}

\abstract{
    Web agents—autonomous systems that navigate and execute tasks on the web on behalf of users—have the potential to transform how people interact with the digital world. However, the most capable web agents today rely on proprietary models with undisclosed training data and recipes, limiting scientific understanding, reproducibility, and community-driven progress. 
We believe agents for the open web should be built in the open. To this end, we introduce (1) \dataset, a large and diverse mixture of browser task demonstrations and web-GUI perception data and (2) \model a family of fully open multimodal web agents.
Specifically, \dataset combines over 100K synthetic task trajectories from multiple complementary generation pipelines with 30K$+$ human demonstrations, atomic web-skill trajectories, and GUI perception data, including referring expression grounding and screenshot question answering. \model agents operate as instruction-conditioned visual-language action policies: given a task instruction and a webpage screenshot, they predict the next browser action, requiring no access to HTML, accessibility trees, or specialized APIs.
Available in 4B and 8B size, on browser-use benchmarks like \webvoyagerfull, \omtwfull, and \deepshop, \model agents achieve state-of-the-art results outperforming similar scale open-weight-only models such as Fara-7B, UI-Tars-1.5-7B, and Holo1-7B. \model-8B also surpasses set-of-marks (SoM) agents built on much larger closed frontier models like GPT-4o. We further demonstrate consistent gains through test-time scaling via parallel rollouts with best-of-N selection, achieving 94.7\% and 60.5\% pass@4 (compared to 78.2\% and 35.3\% pass@1)on \webvoyagerfull and \omtwfull respectively. We will release model checkpoints, training data, code, and a unified evaluation harness to enable reproducibility and accelerate open research on web agents.
}

\begin{document}

\maketitle

\setcounter{tocdepth}{2}


\section{Introduction}
\label{sec:intro}

\begin{figure}[t]
    \centering
    \includegraphics[width=0.8\textwidth]{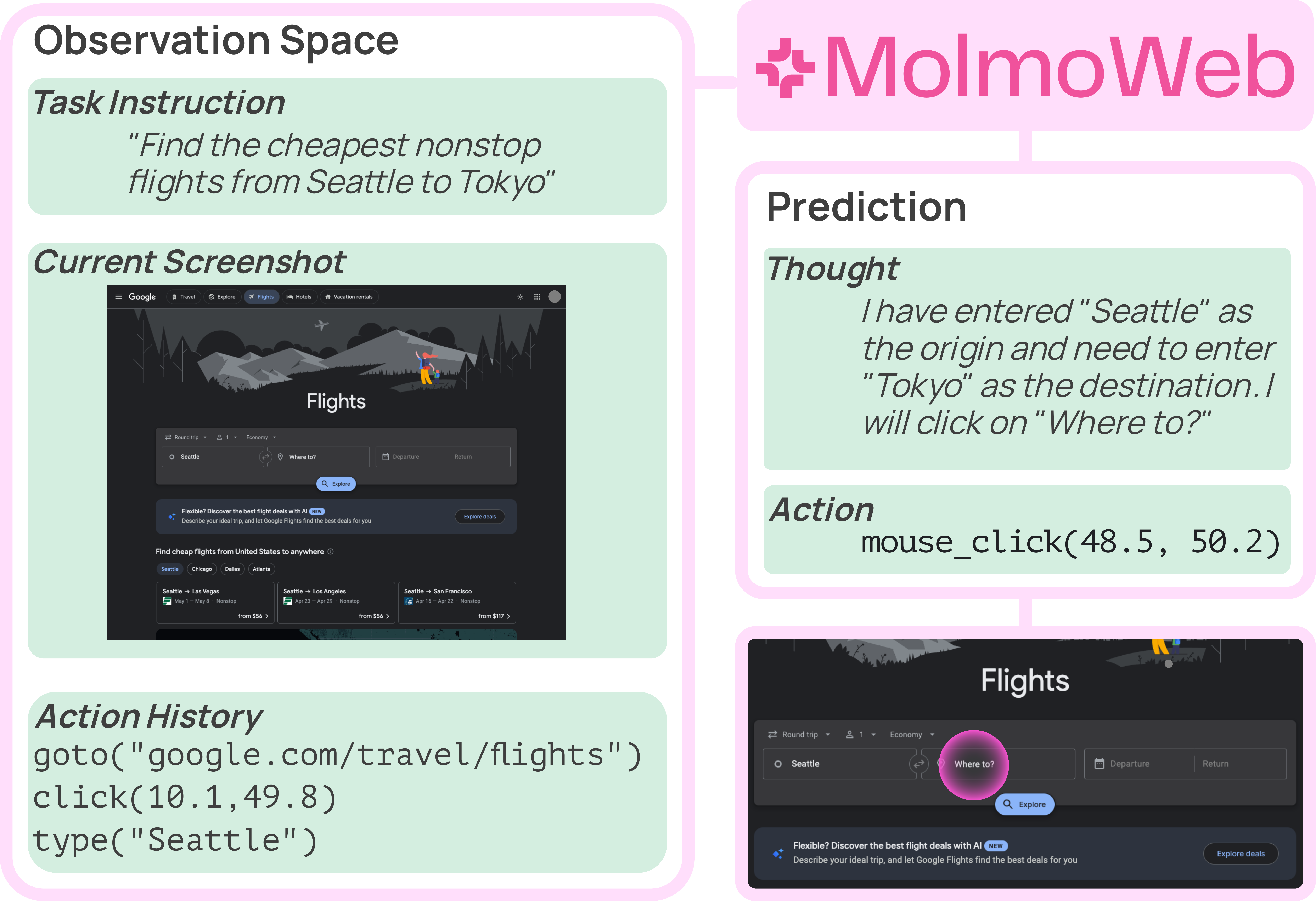}
    \caption{\textbf{Overview of \model.} At each step, the agent receives the task instruction, current screenshot, and action history as input. It then predicts a natural language thought followed by a browser action.
    }
    \label{fig:model}
\end{figure}

\begin{figure}[t]
    \centering
    \includegraphics[width=0.8\textwidth]{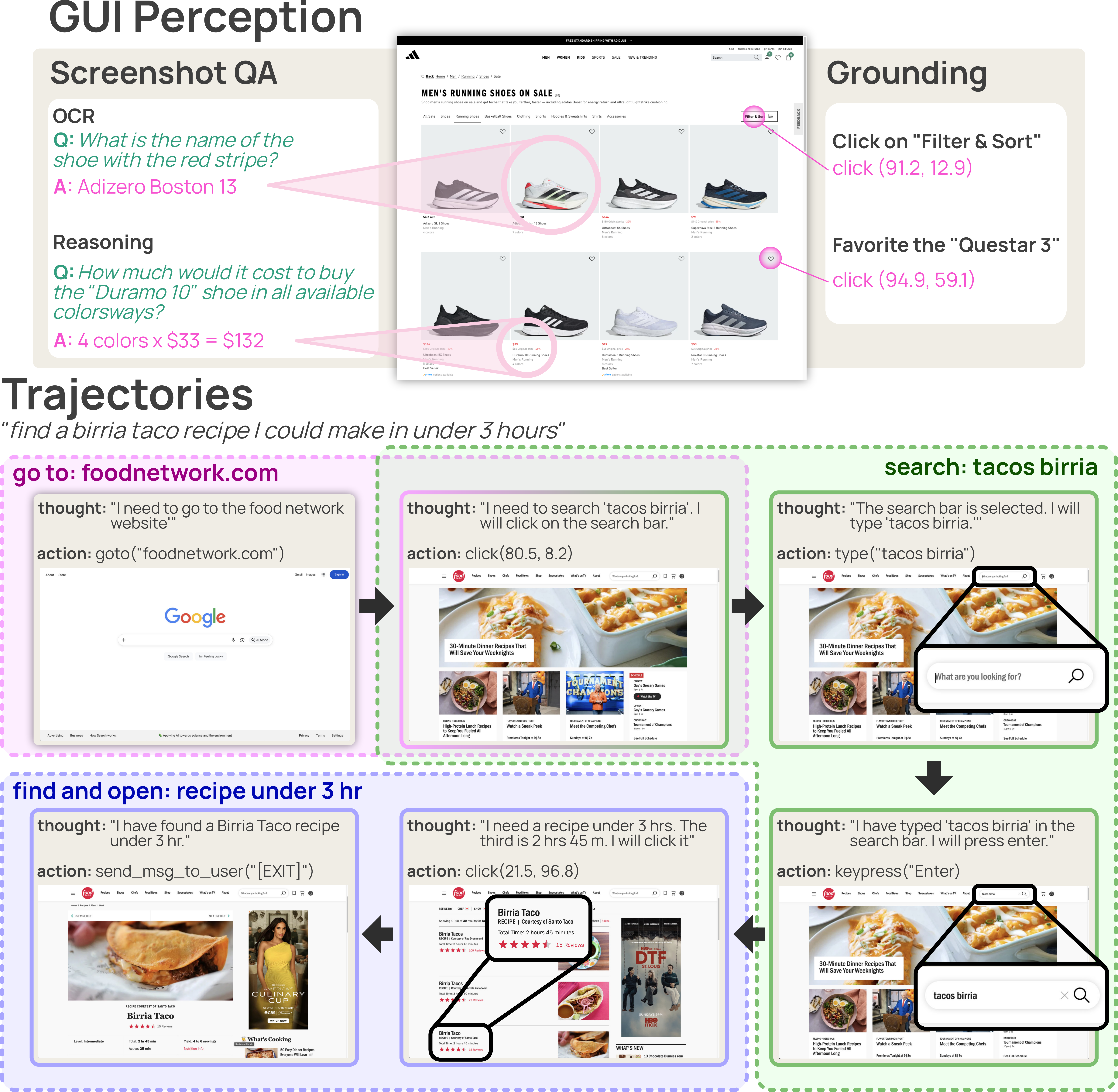}
    \caption{\textbf{Overview of \dataset dataset.}  To capture the diverse capabilities required for web agents, \dataset combines GUI perception data involving screenshot QA \textbf{(top left)} and referring expression grounding \textbf{(top right)}, with synthetic and human task trajectories demonstrating task completion on the web \textbf{(bottom)}.  The human trajectories are further segmented into atomic skills (\Cref{tab:skills} shows our skill taxonomy).}
    \label{fig:teaser}
\end{figure}

The World Wide Web is an indispensable part of modern human life, with billions of people relying on it daily for information, communication, commerce, and entertainment~\cite{itu2024facts}. From booking flights and managing finances to navigating government services, comparison shopping, and coordinating logistics, countless everyday tasks require users to interact with complex, multi-step web interfaces that demand sustained attention, domain knowledge, and patience~\cite{itu2024facts,oecd2023digitalskills}. Web agents---autonomous systems that can navigate and execute such tasks on the web at the behest of users~\cite{nakano2021webgpt}---have the potential to fundamentally reshape how humans interact with this vast digital ecosystem~\cite{who2023disability,w3c2024abilitiesbarriers}. 
Effective web agents could broaden digital access for users who face barriers due to disability or inaccessible design~\cite{who2023disability,w3c2024abilitiesbarriers}, as well as for users with limited digital skills or digital literacy~\cite{oecd2023digitalskills}.

Recent progress in large language and vision-language models has brought web agents closer to practical viability, supported by research benchmarks demonstrating multi-step web interaction (e.g., clicking, form filling, and navigation) under natural-language instructions~\cite{deng2024mind2web,Zhou2023WebArena,shi2017worldofbits,liu2018reinforcement}. 
In parallel, frontier labs now expose computer- or browser-use capabilities where models perceive GUI state via screenshots and output low-level actions such as click/type/scroll to complete workflows~\cite{openai_computeruse_guide,openai_developers_2025,google_gemini_computeruse}. 
However, the most capable end-to-end systems are typically offered as hosted, proprietary services with limited disclosure of training data and full recipes~\cite{openai_computeruse_guide,google_gemini_computeruse}. 
This exacerbates long-standing concerns that insufficient reporting and artifacts hinder reproducibility and scientific understanding of what actually drives performance~\cite{gundersen2018repro,pineau2021repro}. 
The resulting opacity is especially concerning for autonomous agents operating on the open web, where trustworthy deployment depends on transparency, accountability, and auditable behavior~\cite{nist2023airmf,weidinger2022taxonomy}.

\textbf{We believe agents for the open web should be built in the open.} 
To this end, we provide the research community with a complete, transparent, and reproducible foundation including: \textbf{\dataset}, a mixture of diverse task demonstrations on the browser and web-GUI perception data (Sec.~\ref{sec:data}); and \textbf{\model}, a family of fully open and efficient multimodal web agents (Sec.~\ref{sec:model}). \model agents are vision-language models (VLMs) trained to serve as instruction-conditioned multimodal action policies for the web. Given a task instruction, action history, and a screenshot of the webpage, \model agent predicts the next browser action. The action is executed in the browser to yield a new observation and this loop repeats until the task is complete. 

Crucially, \model agents operate on any website using only the visual interface that human users see, without requiring specialized APIs or access to the underlying HTML or accessibility tree (AxTree). This vision-centric design is a deliberate choice motivated by several considerations. First, it mirrors how people actually use the web, grounding the agent in the same perceptual modality and making its behavior more interpretable. 

Second, it avoids the brittleness of Document Object Model (DOM) based representations, which vary significantly across websites, frameworks, and even minor page updates, and which can be incomplete or misleading for dynamically rendered content. Third, it sidesteps the substantial token consumption and compute overhead that structured page representations incur: AxTree inputs can easily consume tens of thousands of tokens per page, whereas a single screenshot provides a compact, information-rich representation of the same content.

The key to training capable \model agents is curating a high-quality and diverse training data mixture that simultaneously teaches the model to understand web screenshots as well as task execution behaviors, such as filling forms, multi-hop navigation, and finding relevant information from a webpage. \dataset contains data from four complementary data sources: (1) synthetic task trajectories generated at scale from multiple complementary pipelines, (2) human demonstrations collected by crowdworkers on real websites, (3) atomic web skill trajectories providing targeted supervision for compositional browser-use skills, and (4) GUI perception data consisting of referring expression grounding and screenshot question-answering examples.

Despite their relatively compact size at 4B or 8B scale, \model agents are highly performant. \textbf{Without distilling from other visual web agents}, \model agents are competitive with or outperform open-weights agents of comparable size, including Fara-7B~\cite{fara7b2025} and Holo1-7B~\cite{Andreux2025Holo1}.
More notably, they also outperform set-of-marks (SoM) prompting based web agents~\cite{yang2023set} built on much larger closed frontier models like GPT-4o that take both AxTree and SoM-annotated screenshots as inputs. 
This result is particularly striking because these closed-model baselines enjoy substantially richer input representations \emph{and} orders-of-magnitude more parameters, yet a well-trained, vision-only \model agent achieves stronger task completion rates---suggesting that data quality and targeted training can compensate for raw model scale and additional inputs.
Finally, we demonstrate further gains by leveraging increased compute at inference-time via multiple rollouts with best outcome selection using a VLM judge.

\section{\dataset}
\label{sec:data}
Our training data consists of 3 broad categories: task trajectories, atomic skill trajectories, and GUI perception data.  The trajectory generation pipeline encompasses web browsing task sampling, trajectory execution, and filtering out failed trajectories, as shown in \Cref{fig:taskgen}.

\subsection{Task trajectories}
\label{subsec:task_trajs}

\begin{figure}[t]
    \centering
    \includegraphics[width=0.99\textwidth]{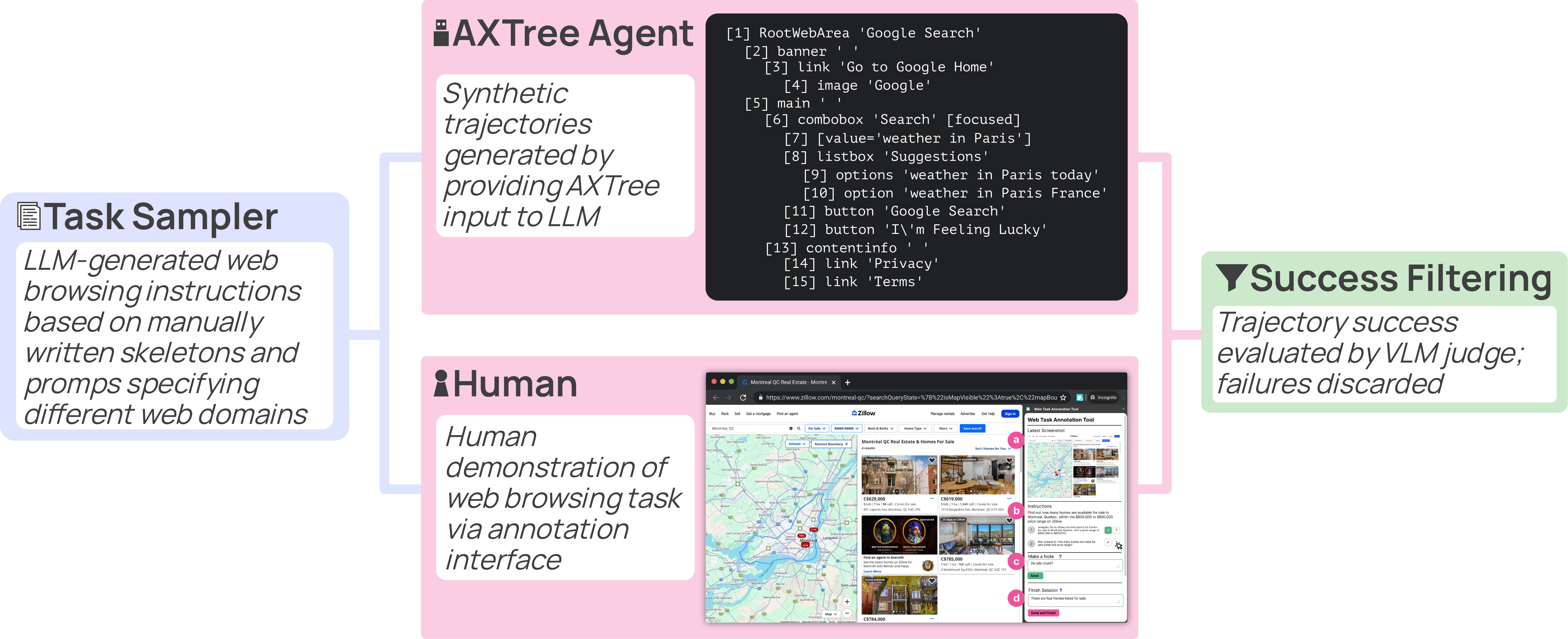}
    \caption{\textbf{Data generation pipeline}. Our pipeline consists of a task sampler, trajectory generation executed by either humans or AxTree agents, and success filtering.}
    \label{fig:taskgen}
\end{figure}

\noindent\textbf{Trajectories from AxTree agents.}
A bulk of our synthetic trajectories are generated using an LLM agent that operates on the AxTree representation of the webpage. At each step, the agent receives the serialized AxTree (visualized in \Cref{fig:taskgen}) as its observation along with the instruction and past actions, and predicts the next action by referencing element browser IDs (\texttt{bid}) assigned to each interactive node. We used Gemini-3-Flash-Preview as the agent backbone.

Tasks were sourced from a combination of manually authored and LLM-generated instructions, covering popular websites and benchmark evaluation sites (see \Cref{sec:supp_synthetic} in the Appendix for details). Although the agent observes only the AxTree, we capture a screenshot at each step so that all trajectories share a unified observation format. Actions predicted in bid-space are post-processed into pixel-space coordinates (clicks, typing, scrolling). See \Cref{tab:actions}
for the complete list of actions.

Synthetic trajectories were filtered for task success using the WebVoyager LLM judge~\cite{He2024WebVoyager}, retaining only trajectories deemed successfully completed.

\begin{figure}[t]
    \centering
    \includegraphics[width=0.8\textwidth]{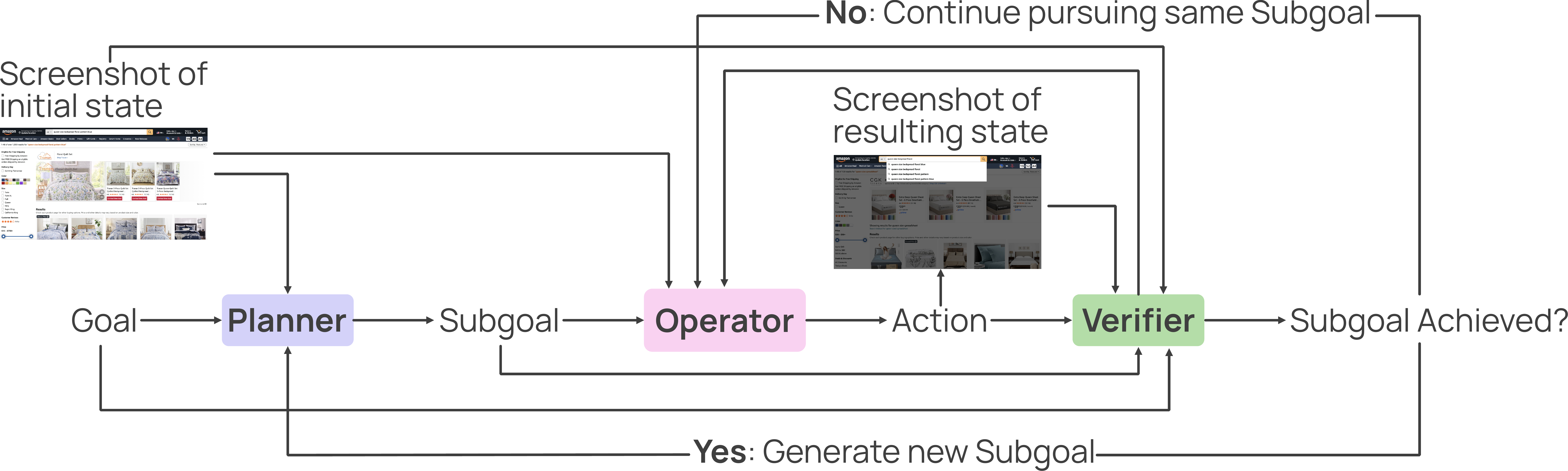}
    \caption{\textbf{Multi-agent trajectory generation pipeline.} A Planner decomposes the task into sequential subgoals. For each subgoal, an Operator generates an action conditioned on the current state and subgoal. After execution, the resulting state is evaluated by a Verifier using screenshots to determine whether the subgoal is achieved. If not, the system continues attempting the same subgoal; otherwise, the Planner generates the next one. The loop repeats until the overall goal is completed.}
    \label{fig:multiagent}
\end{figure}

\noindent\textbf{Trajectories from multiagent harness.}
To generate trajectories with potentially a higher quality of thoughts and trajectories, we design a multi-agent system (\Cref{fig:multiagent}) where the the system orchestrates web navigation through three specialized roles working in concert: a \texttt{Planner} generates the immediate next subgoal based on the high-level task goal as well as task progress determined from verification feedback; an \texttt{Operator} executes one low-level browser action (like clicking, scrolling etc.) at each step to accomplish the current subgoal; and a \texttt{Verifier} checks whether the current subgoal has been completed by analyzing the most recent 5 screenshots. The trajectory generation process operates iteratively. At each step, the system first runs the \texttt{Verifier} to check if the current subgoal is complete. If the subgoal is verified as complete or fails to complete within 5 steps, the \texttt{Planner} is called to generate the next subgoal. The system then injects the current subgoal, completed steps, and planner/verifier reasoning into the text prompt and passes this augmented prompt along with the current screenshot to the \texttt{Operator}, which returns a concrete browser action. Finally, all historical information is tracked and fed back into subsequent planning and execution cycles. We use Gemini-2.5-Flash for the \texttt{Planner}, GPT4-o as the \texttt{Verifier}, and Gemini AxTree agent as the Operator. We empirically find that this multi-agent setup achieves higher task completion success rates than using the Gemini AxTree agent alone, scoring 78.5 vs. 74.4 on \webvoyagerfull~\cite{He2024WebVoyager}. 

\noindent\textbf{Trajectories from humans.}
In addition to the synthetic trajectories, we collected human demonstrations via a custom Chrome extension that captures browser interaction events (e.g., clicks, scrolls, keystrokes) and corresponding screenshots, and post-processes them into standardized trajectories. Crowdworkers were provided with tasks sourced from a mix of manually authored and LLM-generated instructions (see \Cref{sec:supp_human} in the Appendix for details). In general, tasks were chosen to span a wide range of popular websites and to require a variety of browser task-completion skills including search, form-filling, and multi-hop navigation. 

To support fine-grained annotation, each task instruction was decomposed into an ordered sequence of subtasks. Workers checked off each subtask upon completion within the annotation tool and submitted a final text response (e.g., an answer to a question or a completion acknowledgment). If a subtask could not be completed due to missing content or an unexpected webpage state, workers recorded their best attempt along with a descriptive note explaining the failure.

Each trajectory is reviewed by a human to verify task completion and ensure that screenshots and actions were correctly captured. Trajectories failing review were revised or re-collected.

\begin{figure}[t]
    \centering
    \includegraphics[width=0.9\textwidth]{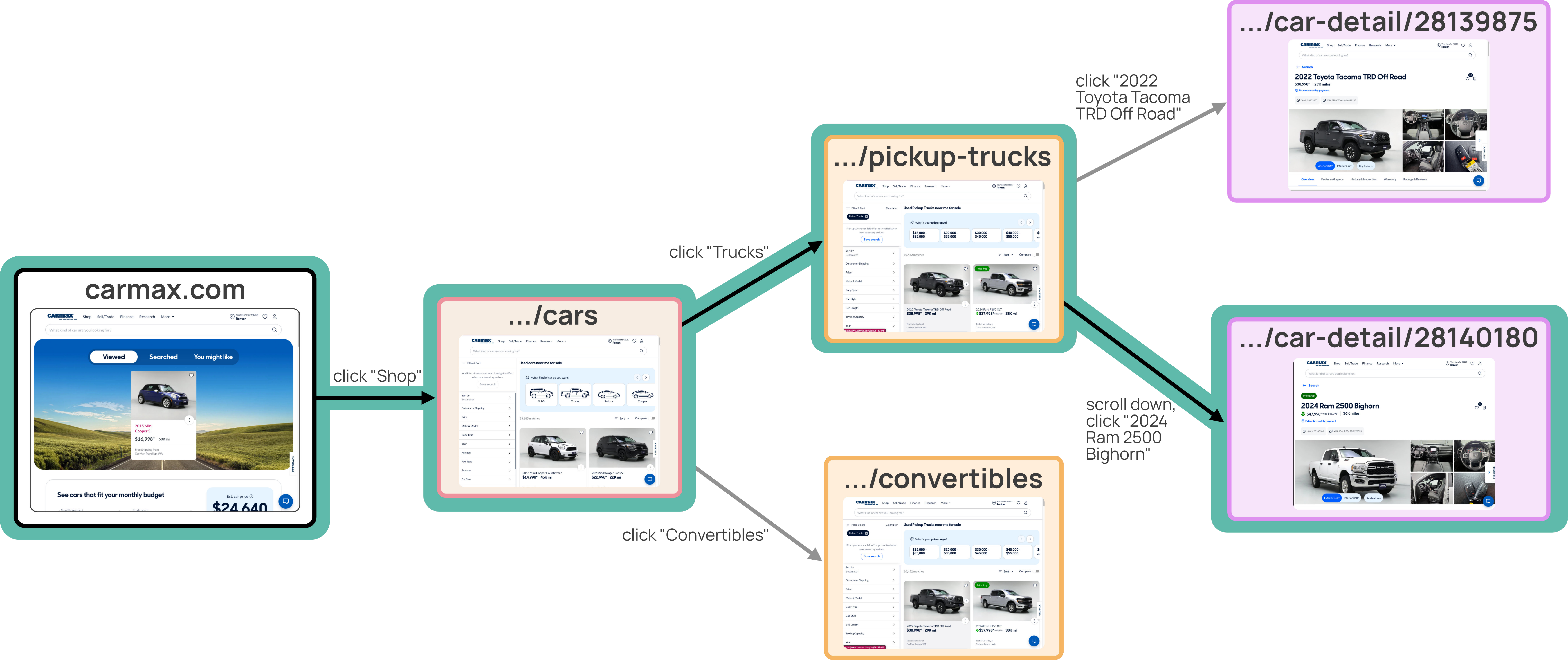}
\caption{\textbf{Node traversal trajectory generation pipeline.} Starting from a website's homepage, we extract all reachable URLs and use an LLM to select the most informative ones for breadth-first expansion, pruning cycles to avoid loops. A deterministic, LLM-free procedure then generates trajectories by traversing the resulting paths via scrolling and clicking, with success verified through URL matching. Each validated trajectory is paired with an LLM-generated goal conditioned on the full trajectory, producing a task consistent with the produced path.}
    \label{fig:nodetrav}
\end{figure}

\noindent\textbf{Trajectories from node traversal.}
To complement LLM and human generated task-completion trajectories with verifiable, deterministic supervision, we construct navigation trajectories from precomputed website graphs. We first build a directed graph over 500 popular websites via breadth-first exploration: starting from the homepage, we extract the accessibility tree at each page and prompt an LLM to select a diverse set of navigational links (e.g., category pages, search features, content sections), continuing to a depth of four. Root-to-leaf paths through each graph yield URL sequences that serve as ground-truth navigation plans, as shown in \Cref{fig:nodetrav}.

To convert a chosen sequence of URLs in the node-graph to a trajectory, we use a deterministic process that does not rely on an LLM to replay each path using browser actions like clicking and scrolling.  Starting at the root URL, we locates the link to the next target in the accessibility tree, scroll if necessary to bring it into view, and click. Because the agent uses no language model at execution time, trajectories are cheap to produce at scale, and the intended route is known \emph{a priori}, enabling straightforward success verification. When the agent cannot reach a target URL, the path is truncated to the last successfully visited page. At the terminal page of each trajectory, we use an LLM to generate a plausible instruction. Together the instruction-trajectory pair generated via a walk on the node-graph now resembles a goal-directed browser demonstration. 

\subsection{Atomic skill trajectories}
\label{subsec:skill_trajs}

While task trajectories require the agent to compose multiple web skills to complete an end-to-end goal (like searching for a product, applying filters, and adding to cart; each requiring a sequence of low-level browser actions), atomic skill trajectories isolate individual skills. We identify a core taxonomy of web skills that underlie most browsing tasks, including search, form filling, finding an element or information on page, and filtering (see \Cref{tab:skills}). Providing targeted supervision for each skill ensures the model develops reliable competence in these building blocks.

\begin{table}[t]
\centering
\caption{\textbf{Taxonomy of atomic web skills used in human annotation.}}
\label{tab:skills}
\small
\setlength{\tabcolsep}{4pt}
\renewcommand{\arraystretch}{1.15}
\begin{tabular}{p{0.38\linewidth}p{0.54\linewidth}}
\toprule
\textbf{Skill} & \textbf{Description} \\
\midrule
\texttt{go\_to}                    & Navigate directly to a specified URL \\
\texttt{search}                    & Enter a query into a search box and submit \\
\texttt{find}                      & Locate an element or information on the page \\
\texttt{find\_and\_open}           & Locate an element or sub\-page and open it \\
\texttt{find\_and\_click}          & Locate an element and click it \\
\texttt{fill\_form}                & Fill form fields with specified values \\
\texttt{fill\_form\_and\_submit}   & Fill form fields and submit the form \\
\texttt{apply\_filters}            & Set filter or sort controls to specified values \\
\texttt{apply\_filters\_and\_search} & Set filters and trigger a search \\
\texttt{add\_to\_cart}             & Add the current item to a shopping cart \\
\texttt{navigate}                  & Free-form navigation when stepwise decomposition is unknown \\
\bottomrule
\end{tabular}
\end{table}

\noindent\textbf{Extracted from human trajectories.}
Because human task trajectories were annotated with ordered subtask decompositions, each subtask segment naturally constitutes an atomic skill trajectory. Specifically, each segment begins from the browser state in which the previous subtask ended, and carries the subtask instruction as its goal. We automatically extract these segments from the full human trajectories, yielding short, focused demonstrations. Since skill segments within a trajectory begin from the browser state where the last skill segment ends, these demonstrations resemble response to a follow-up query to the user's previous query. Therefore, this data might also enable the agent to learn multi-turn interaction with the user. 

\noindent\textbf{Generated by an AxTree agent.}
To supplement the extracted segments, we additionally prompted the AxTree agent to execute targeted skill instructions for two important skills: \texttt{fill\_form} and \texttt{find\_and\_open}. For each, instructions were of the form ``\texttt{go to:[URL]\textbackslash{}n fill form:[form details]}'' or ``\texttt{go to:[URL]\textbackslash{}n find and open:[target]}.'' This directly yields skill-level trajectories without requiring segmentation of longer task trajectories.

\begin{table}[t]
\centering
\small
\caption{\textbf{Statistics of \dataset}. We also report its constituent sources compared to other datasets.}
\setlength{\tabcolsep}{2pt}
\renewcommand{\arraystretch}{1.15}
\begin{tabular}{lcccccc}
\toprule
\textbf{Dataset / Source} & \textbf{Trajectories} & \textbf{Steps} & \textbf{Domains} & \textbf{Avg steps} & \textbf{Mixture Ratio} & \textbf{Open-source} \\
\midrule
Mind2Web~\cite{deng2024mind2web} & 2.4K & 17K & 137 & 7.3 & -- & Yes \\
AutoWebWorld~\cite{wu2026autowebworldsynthesizinginfiniteverifiable} & 11.6K & 255K & 29 & 21.9 & -- & Yes \\
Fara~\cite{fara7b2025} & 145K & 1.01M & -- & 6.9 & -- & No \\
\rowcolor{gray!15}
\textbf{MolmoWebMix-Traj} & 278.5K & 2.2M & 2.6K & 13.2 & 0.80 & Yes \\
\rowcolor{gray!15}
\multicolumn{7}{l}{\textit{Synthetic}} \\
\rowcolor{gray!15}
\quad AxTree Single-Agent & 70K & 793K & 1.3K & 11.4 & 0.35 & Yes \\
\rowcolor{gray!15}
\quad Axtree Multi-Agent & 35K & 438K & 1.1K & 12.5 & 0.18 & Yes \\
\rowcolor{gray!15}
\quad Axtree Atomic Skills & 5.5K & 68.7K & 540 & 12.4 & 0.02 & Yes \\
\rowcolor{gray!15}
\quad Node-Traversal & 16K & 151K & 833 & 9.5 & 0.02 & Yes \\
\rowcolor{gray!15}
\multicolumn{7}{l}{\textit{Human}} \\
\rowcolor{gray!15}
\quad Human & 36K & 623K & 1.1K & 20.8 & 0.18 & Yes \\
\rowcolor{gray!15}
\quad Human Skills & 116K & 781K & 1.1K & 6.8 & 0.05 & Yes \\
\rowcolor{gray!15}
\textbf{MolmoWebMix-Perception} & -- & 10.5M & -- & -- & 0.20 & Yes \\
\rowcolor{gray!15}
\quad PixMoPoints + SyntheticGround & -- & 8.3M & -- & -- & 0.15 & Yes \\
\rowcolor{gray!15}
\quad ScreenshotQA & -- & 2.2M & -- & -- & 0.05 & Yes \\
\bottomrule
\end{tabular}
\label{tab:traj_stats}
\end{table}

\subsection{GUI perception}
\label{subsec:gui_perception}
Effective web navigation requires the agent to perceive and understand webpage screenshots before taking any action. GUI perception data teaches the model to identify interactive elements on the page (e.g., buttons, links, input fields, and other controls), understand their function and affordances, and extract information from the page in response to natural language queries. This combines skills akin to referring expression grounding, OCR, and reading comprehension, grounded in the visual structure of web interfaces.

\noindent\textbf{Grounding.}
Grounding data consists of (screenshot, element description) $\to$ click point pairs, where the model must predict the pixel coordinate to click for interacting with a specified element. We extract these pairs automatically from AxTree agent trajectories. For each screenshot, we enumerate all clickable elements in the AxTree and for each generate a natural language description using its accessible name and role either with a template or GPT5. The ground-truth click coordinate is sampled randomly within the element's bounding box using a clipped Gaussian prior centered at the element's center, encouraging the model to learn spatially robust clicking rather than always targeting exact centers. In total, we generate more than 7M grounding QA pairs, comprising 3.4M templated queries and 3.8M GPT-5-generated queries. 
In addition to new grounding data, we also repurpose the PixmoPoints data from Molmo~\cite{clark2026molmo2openweightsdata} by formatting single-point QA pairs into click actions and include 1.1M examples in our grounding mixture. 

\noindent\textbf{Screenshot QA.}
Screenshot QA data consists of (screenshot, question) $\to$ answer pairs that enhance the model's ability to read and reason about webpage content. For each screenshot in a subset of the AxTree agent trajectories, we provide the corresponding AxTree to an LLM and prompt it to generate question--answer pairs. Questions cover three categories: OCR queries about text and values present on the page (e.g., prices, counts, text content), affordance queries about actions available on the page (e.g., ``Where would I find financial news on this page?''), and summarization queries about the overall content or purpose of a page element. To ensure that questions and answers rely solely on visual content, we remove samples that contain references to AxTree-specific information, such as element IDs (e.g., ``Click on Bid 32''). In total, the dataset covers 395 websites and 2,237,252 QA pairs, split between 54\% OCR, 26\% affordance, and 20\% summarization questions.

\section{\model}
\label{sec:model}
In this section, we describe the \model architecture, its observation and action space, and the training procedure. \model is designed to be simple and practical: we build on an existing vision-language model backbone, define a minimal but expressive action space over browser operations, and train end-to-end via supervised fine-tuning on the \dataset mixture.

\subsection{Architecture}
\label{subsec:architecture}
\model is built on the Molmo2 architecture~\cite{clark2026molmo2openweightsdata}, a multimodal language model that processes interleaved sequences of images and text. As shown in \Cref{fig:model}, the model takes as input the current webpage screenshot together with the task instruction and the history of past actions, and produces a structured output comprising a natural language thought followed by the next action to execute.

\subsection{Observation and action space}
\label{subsec:obs_action}

\noindent\textbf{Observations.} 
At each step $t$, the model receives a screenshot of the current browser viewport along with the task instruction. To provide temporal context, the history of actions taken in 10 prior steps is appended as context along with the URL and title of the current page.

\noindent\textbf{Actions.} The model predicts a JSON object encoding both a \emph{thought} and an \emph{action} specifying the operation to perform. The thought is a brief natural language rationale for the chosen action. The full action space is described in \cref{tab:actions}. Mouse actions are parameterized by spatial coordinates normalized to $[0, 100]$, with 2 decimal points, which are denormalized to viewport pixel coordinates at execution time.

\begin{table}[t]
\centering
\caption{\textbf{Action space of \model.} Spatial coordinates are normalized to $[0, 100]$ during training and denormalized to viewport pixels for browser execution. \Cref{fig:action_dist} in the Appendix shows the distribution of actions in various subsets of \dataset.}
\label{tab:actions}
\setlength{\tabcolsep}{4pt}
\renewcommand{\arraystretch}{1.15}
\small
\begin{tabular}{p{0.42\linewidth}p{0.50\linewidth}}
\toprule
\textbf{Action} & \textbf{Description} \\
\midrule
\texttt{goto(url)}                         & Navigate the browser to a URL \\
\texttt{mouse\_click(x, y, ...)}           & Click at a viewport coordinate \\
\texttt{mouse\_drag\_and\_drop(...)}       & Drag from one coordinate to another \\
\texttt{scroll(delta\_x, delta\_y)}        & Scroll the page by an offset \\
\texttt{scroll\_at(x, y, dx, dy)}          & Scroll at a specific coordinate \\
\texttt{hover\_at(x, y)}                   & Hover at a specific coordinate \\
\texttt{keyboard\_type(text)}              & Type a string of text \\
\texttt{keyboard\_press(key)}              & Press a key or key combination \\
\texttt{go\_back()}                        & Navigate to the previous page \\
\texttt{new\_tab()}                        & Open a new browser tab \\
\texttt{tab\_focus(index)}                 & Switch focus to a browser tab \\
\texttt{noop(wait\_ms)}                    & Wait (e.g., for page load or captcha) \\
\texttt{send\_msg\_to\_user(msg)}          & Display a message to the user \\
\bottomrule
\end{tabular}
\end{table}

\subsection{Training}
\label{subsec:training}

We train \model end-to-end via supervised fine-tuning (SFT) on all training data described in \cref{sec:data}. All data types, including task trajectories, atomic skill trajectories, and GUI perception data, are mixed in a single training stage. The mixing ratios across data sources are treated as hyperparameters; we ablate different mixtures and select the combination that best balances performance across evaluation benchmarks. We report the datasets' final ratios in the mixture in Table \ref{tab:traj_stats}.

We finetune Molmo2~\cite{clark2026molmo2openweightsdata} (with Qwen3 language model and SigLIP2 vision encoder) following its best practice and tune the language model, the vision encoder, and the adapter starting from the single-image checkpoint (pretrained on image captioning and finetuned on single-image QA). We train all models with 64 H100 GPUs with a global batch size of 128 for up to 50K steps (approximately 3.2 epochs on average, with slightly different numbers of epochs across datasets based on their final ratio in the mixture).
\section{Experiments}
\label{sec:exps}

Our experiments consist of comparison to prior art, a study on scaling test-time compute through more inference steps per rollout as well as parallel rollouts with best-of-N selection, data ablation studies to understand the impact of scale and role of human vs synthetic data, comparison of token sampling strategies, and grounding evaluation. We first describe our evaluation setup and then present our experimental results.

\begin{table}[t]
\caption{\textbf{Comparison to prior work on browser-use benchmarks.} For models marked with a *, the numbers reported are from Fara~\cite{fara7b2025}. Italic \omtwfull results are from the Online-Mind2Web auto-eval leaderboard.  Since its unclear what judge was used by \cite{fara7b2025} for \webtailfull we used the \webvoyagerfull judge to compute numbers marked with \dagger. We highlight the \best{best} and \sbest{second-best} performing numbers for each benchmark in the sub-8B, open-weight model category.}
\centering
\scalebox{0.8}{
\setlength{\tabcolsep}{4pt}
\renewcommand{\arraystretch}{1.15}
\begin{tabular}{lccccc}
\toprule
Model & \# Steps & \webvoyagerfull~\cite{He2024WebVoyager} & \omtwfull~\cite{Xue2025Om2w} & \deepshop~\cite{Lyu2025DeepShop} & \webtailfull~\cite{fara7b2025} \\
\midrule
\multicolumn{6}{l}{\textbf{API only}} \\
\multicolumn{6}{l}{\textit{w/o vision}} \\
\quad Axtree Agent (GPT-5) & 30 & 70.6 & 41.9 & 40.7 & 29.2$^\dagger$ \\
\quad Axtree Agent (Gemini-3-flash)  & 30 & 74.4 & 34.8 & 45.1 & 62.1$^\dagger$ \\
\quad Axtree Agent (Gemini-3-flash) & 100 & 85.6 & 44.8 & 55.3 & 63.5$^\dagger$ \\
\multicolumn{6}{l}{\textit{w/ vision}} \\
\quad SoM Agent (GPT-4o)* & 100 & 65.1 & 34.6 & 16.0 & 30.8 \\
\quad SoM Agent (o3)* & 100 & 79.3 & 55.4 & 49.7 & 52.7 \\
\quad SoM Agent (GPT-5)* & 100 & 90.6 & 57.7 & 49.1 & 60.4 \\

\quad OpenAI computer-use-preview* & 100 & 70.9 & \textit{58.3} & 24.7 & 25.7 \\
\quad Gemini computer-use-preview & 100 & 88.6 & \textit{57.3} & 62.0 & 63.0$^\dagger$ \\
\quad Yutori Navigator~\cite{yutori2025navigator} & -- & -- & \textit{64.7} & -- & -- \\

\hline
\multicolumn{6}{l}{\textbf{Open weight}} \\
Holo1-7B~\cite{Andreux2025Holo1} & 30 & 55.4 & -- & -- & -- \\
UI-TARS-1.5-7B*~\cite{UITARS15} & 100 & 66.4 & 31.3 & 11.6 & 19.5 \\
GLM-4.1V-9B-Thinking*~\cite{hong2025glm} & 100 & 66.8 & 33.9 & 32.0 & 22.4 \\
Fara-7B*~\cite{fara7b2025} & 100 & 73.5 & \sbest{34.1} & 26.2 & 38.4 \\

\hline
\multicolumn{6}{l}{\textbf{Ours: Open weight, data, training, and evaluation}} \\
\model-4B & 100 & \sbest{75.2} & 31.3 & \sbest{35.6} & \sbest{43.8}$^\dagger$ \\
\model-8B & 100 & \best{78.2} & \best{35    .3} & \best{42.3} & \best{49.5}$^\dagger$ \\
\bottomrule
\end{tabular}
}
\label{tab:sota_comparison}
\end{table}

\subsection{Evaluation setup}
We evaluate \model on four popular web browsing benchmarks using live websites: \webvoyagerfull~\cite{He2024WebVoyager}, \omtwfull~\cite{Xue2025Om2w}, \deepshop~\cite{Lyu2025DeepShop}, and \webtailfull~\cite{fara7b2025}. Because some tasks are time-sensitive, we change dates in outdated requests (\eg, find a flight on August 5, 2025) to be meaningful for the task across all benchmarks. For each benchmark and model, we run 3-5 evaluations up to 100 steps and report the average score across runs. If a model does not complete the task by the maximum number of steps, it is considered a failure. As environment errors occasionally occur, we allow up to 10 retries per trajectory; tasks that do not complete within this budget are also marked as failures.  We use the same prompt and LLM-as-a-judge setup corresponding to each published benchmark. Specifically, we use GPT-4o with the official prompt from \webvoyagerfull and \deepshop, and o4-mini with respective judge for \omtwfull. Since, it is unclear what judge was used for \webtailfull, we used the \webvoyagerfull judge. For \webvoyagerfull, we use the task set provided by~\cite{fara7b2025} which removed infeasible tasks. All evaluations are run in a Browserbase environment, allowing many live browsers to run in parallel with some captcha-solving capability.

\subsection{Comparison to prior work}
\label{sec:main_res}
In Tab.~\ref{tab:sota_comparison}, we compare \model to web-agents based on proprietary APIs as well as open-weight models. 

\paragraph{\model establishes a new state-of-the-art among open-weight models.} \model-8B shows healthy gains over leading open-weight models like Fara across all benchmarks. Even \model-4B outperforms all open-weight models on \webvoyagerfull and \deepshop while being competitive on \omtwfull and \webtailfull. These gains largely demonstrate efficacy of \dataset in teaching a strong foundation vision-language model like \basemodel to follow instructions for browser-task completion.   

\paragraph{Comparison to closed visual agents.} Compared to SoM agents that use set-of-marks annotated visual prompts, \model-8B easily outperforms GPT-4o, matches the performance of o3 on \webvoyagerfull, and is only 6 pts behind GPT-5 and o3 on \deepshop. \model-8B also outperforms OpenAI computer-use-preview on \webvoyagerfull and \deepshop as per the performance reported in \cite{fara7b2025}.

\paragraph{Comparison to non-visual teacher agent.} Our synthetic data generation pipeline is powered by Gemini-3-flash axtree agent. It is natural to ask how does the visual student, \model, fair against the non-visual axtree-guided teacher? \model-8B agent is $5+$ pts behind the teacher on \webvoyagerfull and \omtwfull while being $10+$ pts behind on \deepshop and \webtailfull. Besides the observation space, other factors contributing to this gap include: (i) differences in model sizes with Gemini presumably being a much larger model compared to our humble 4B and 8B models); (ii) click action requires pixel-space pointing capability from \model but the Axtree agent uses unique numeric identifiers (\verb|bid|~\cite{Chezelles2024BrowserGym}) of the target element which is then programmatically mapped to a click location; and (iii) answering questions based on text rendered on the webpage requires \model to implicitly perform OCR to parse text from screenshot in addition to reading comprehension required by the Axtree agent.

\subsection{Test-time scaling}
\label{sec:test-time}

\begin{figure}[t]
    \centering
    \includegraphics[width=\linewidth]{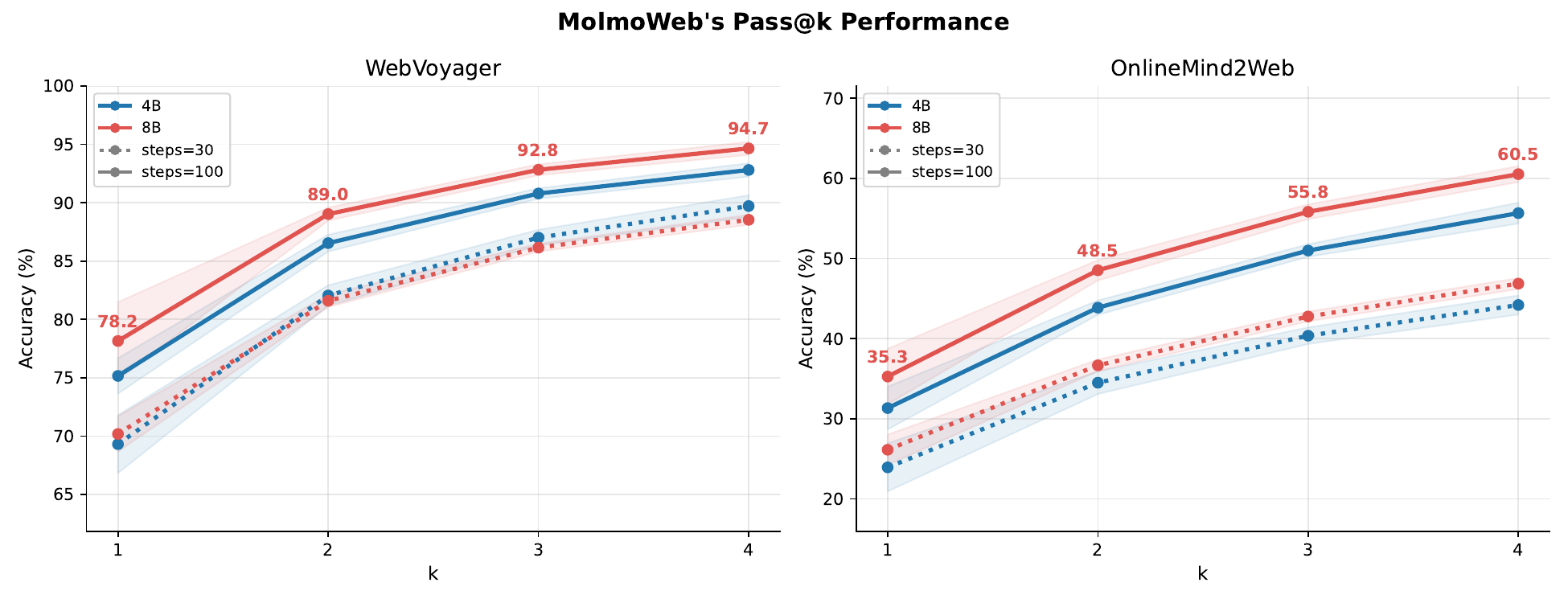}
    \caption{\textbf{Pass@k performance on \webvoyagerfull and \omtwfull} Accuracy improves as k increases, with \model-8B generally outperforming \model-4B, and 100 steps outperforming 30 steps by a large margin. The \model-8B model with max 100 steps reaches 94.7\% and 60.5\% at Pass@4.}
    \label{fig:pass_at_k}
\end{figure}

We explore two ways of utilizing more compute at inference time: (i) performing parallel rollouts with a best-of-N selection with an LLM-judge; and (ii) increasing the number of inference steps.

\paragraph{Parallel rollouts.} We study how much performance can be gained by running $k$ independent agents on the same task and selecting the best outcome.
To get an unbiased, low variance estimate of pass@$k$, we collect $m > k$ rollouts per task and compute the estimate:
\begin{equation}
    \widehat{\text{pass@}k} = 1 - \frac{\binom{m - c}{k}}{\binom{m}{k}},
\end{equation}
where $c$ is the number of successful rollouts among the $m$ attempts. We use an LLM judge to evaluate success for each rollout, and we set $m=5$ to balance accuracy with computational cost. We plot \model-4B and \model-8B's pass@k performance on the \webvoyagerfull benchmark with max steps of 30 and $k=1,...,4$ in Fig.~\ref{fig:pass_at_k} using the respective judge for best rollout selection. We see that for both 4B and 8B models, parallel rollouts improve success rates substantially with pass@$4$ leading to 20+ pt gains over single rollout performance. Parallel rollouts help mitigate the compounding error problem where a single incorrect action could throw the agent off-course. These massive gains from parallel rollouts suggest (or, perhaps explain to the unsurprised why) self-distillation from best-of-N rollouts and RL might be effective strategies for further improving single rollout performance. 

\paragraph{Increasing inference steps.} Figure~\ref{fig:pass_at_k} shows consistent gain from increasing the maximum number of inference steps from 30 to 100 across all benchmarks. However, we note that the gains from scaling via parallel runs (each with 30 steps) and picking the best result using an LLM judge far outperform increasing the number of inference steps (e.g., 8B model achieves $86.2\%$ via 3 parallel runs with max 30 steps resulting in a total of 90 steps compared to $78.2\%$ via increasing the steps to 100 in one run).

\subsection{Training data ablations}
\label{sec:effect_data_scale}

In \Cref{tab:data_ablation}, we evaluate the effect of data scale and contributions from synthetic and human data to final performance. The ablations used a slightly earlier version of \dataset which had fewer human trajectories as well as fewer synthetic trajectories but largely had a similar composition to the final version of \dataset. 

\paragraph{Performance improves with training data scale.} \Cref{tab:data_scale} shows that, unsurprisingly, performance improves with data scale. Interestingly, about $85$ to $90\%$ of the performance can be achieved with just $10\%$ of the dataset highlighting the effectiveness of the training mixture.

\paragraph{Limited gains from human data on benchmarks.} \Cref{tab:effect_human_data} shows that gains from human data are somewhat limited and not consistent across benchmarks. This is likely due to lower volume of human data, differences in task distribution compared to benchmarks, difficulty in learning new actions present only in human data (\texttt{scroll\_at} and \texttt{mouse\_drag\_and\_drop}), difference in style and quality of post-hoc generated thoughts, and human annotation noise (e.g. variable scroll amounts in human annotation as opposed to always scrolling by 100\% of viewport height in synthetic data). We hypothesize that synthetic and human data represent distinctly different web task completion policies and the model struggles to learn generalizable behavior across both. For instance, while we observe the model trained only on human data produces both \texttt{scroll} and \texttt{scroll\_at} actions, when trained along with synthetic trajectories the model almost always produces \texttt{scroll} which tries to scroll the page instead of the element.

\begin{table}[t]
\centering
\caption{\textbf{Data ablations}. These ablations use an earlier version of \dataset with both fewer synthetic and human trajectories compared to the final experiments and use 30 inference steps.}
\label{tab:data_ablation}
\begin{subtable}{0.48\textwidth}
\centering
\caption{Effect of training data scale}
\label{tab:data_scale}
\begin{tabular}{lcc}
\toprule
Training data & \webvoyager & \omtw \\
\midrule
1\% & 44.5 & 11.7 \\
10\% & 63.2 & 20.4 \\
100\% & \textbf{68.5} & \textbf{21.9} \\
\bottomrule
\end{tabular}
\end{subtable}
\hfill
\begin{subtable}{0.48\textwidth}
\centering
\caption{Humans vs. synthetic data}
\label{tab:effect_human_data}
\begin{tabular}{lccc}
\toprule
Training data & \#trajs & \webvoyager & \omtw \\
\midrule
human only     & 28K & 27.8 & 13.2 \\
synthetic only & 106K & 67.8 & \textbf{22.0} \\
synthetic $+$ human & 134K & \textbf{68.5} & 21.4 \\
\bottomrule
\end{tabular}
\end{subtable}

\end{table}

\paragraph{Synthetic trajectories are more effective for learning than human trajectories for the same tasks.} To further investigate the difference in learning from trajectories collected by AxTree agents and humans, we collect 2700 web browsing trajectories with each method, using the same task instructions. We then train the model on each of these datasets, mixed with \dataset-Perception data. \Cref{tab:llm_vs_human} shows that the model trained on AxTree-sourced trajectories consistently outperforms the one trained on human trajectories, suggesting that the synthetic data provides a more reliable learning signal. We hypothesize two contributing factors for higher signal-to-noise ratio in synthetic trajectories. First, humans tend to exhibit more exploratory behavior, particularly on unfamiliar websites, resulting in longer and noisier trajectories with detours that may hinder imitation learning. Second, the LLM agent operates on the accessibility tree, which encodes rich structural and semantic information about page elements that may not be immediately apparent from visual cues alone. This additional context likely enables the LLM to produce more direct and consistent trajectories, making them more effective as training data despite being synthetically generated.
\begin{table}[h]
\centering
\caption{\textbf{Learning from LLM-generated trajectories is more effective than learning from human demonstrations.}}
\label{tab:llm_vs_human}
\begin{tabular}{lcccc}
\toprule
Source & \#trajs &\deepshop & \webvoyager & \omtw \\
\midrule
human & 2.7K &19.8 & 35.4 & 9.0 \\
synthetic   & 2.7K & \textbf{24.4} & \textbf{53.0} & \textbf{16.8} \\
\bottomrule
\end{tabular}
\end{table}

\subsection{Sampling Strategies}
\label{sec:sampling_strategies}

\Cref{tab:sampling_strategies} compares greedy, top-k, and top-p (nucleus) sampling on \webvoyager. We find that sampling strategy significantly affects \model{}'s performance, achieving over 5\% gains with top-k and top-p sampling compared to greedy decoding. Qualitatively, we find the greedy-decoding could get stuck at specific states (e.g. trying to click at the same location or continuing to scroll even when past attempts at doing so have failed), while the other randomized strategies overcome this behavior. Between top-k and top-p sampling, we see that nucleus sampling with p=0.8 and temperature=0.7 performs the best on Webvoyager and hence is the default sampling strategy used for all experiments. We choose the sampling parameters based on Qwen3's (the base LLM used in \basemodel) recommended parameters on HuggingFace~\cite{wang2024qwen2vl}.  

\begin{table}[h]
  \caption{\textbf{Effect of sampling strategy.} Greedy sampling performs significantly worse than random sampling strategies.}
  \label{tab:sampling_strategies}
  \centering
  \begin{tabular}{lc}
    \toprule
    Strategy & \webvoyager \\
    \midrule
    greedy (temperature=0.0) & 61.4 \\
    top-k (temperature=0.7, k=20) & 67.4 \\
     top-p  (temperature=0.7, p=0.8) & \textbf{68.5} \\
    \bottomrule
  \end{tabular}
\end{table}

\subsection{Grounding Experiments}

\begin{table}[t]
\centering
\small
\caption{We present \model{}-4B's and \model{}-Ground's (grounding specialist trained on grounding data only) performance on two prominent benchmarks: ScreenSpot and ScreenSpotV2 against open and closed baselines. \best{Best} and \sbest{second-best} numbers are highlighted.}
\begin{tabular}{lcc}
\hline
Model & ScreenSpot & ScreenSpot v2 \\
\hline
Claude 3.7~\cite{anthropic2024claude} &  --- & 87.6 \\
OpenAI CUA~\cite{openai_computeruse_guide}&  --- & 87.9 \\
Gemini-3-Pro~\cite{gemini3} &  --- & \best{93.7} \\
\hline
UGround-7B~\cite{Qian2025UGroundTU}&  86.7 & 76.3 \\
Qwen2.5VL-7B~\cite{wang2024qwen2vl} &  85.5 & 89.3 \\
Holo1-7B~\cite{Andreux2025Holo1} & \sbest{87.4} & 89.9 \\
Fara-7B~\cite{fara7b2025} & 86.7 & 89.3 \\
\hline
\model{}-4B & 87.2 & 89.5 \\
\model{}-Ground-8B & \best{88.7} & \sbest{91.8}\\
\hline
\end{tabular}

\label{tab:grounding}
\end{table}
In \Cref{tab:grounding}, we evaluate \model{}-Ground-8B---an 8B grounding specialist trained only on grounding data as well as the \model{}-4B agent on two grounding benchmarks: ScreenSpot~\cite{Cheng2024SeeClick} and ScreenSpot v2~\cite{wu2024atlas}. We see that the grounding specialist outperforms open-weight models as well as much larger proprietary models like Claude 3.7 and OpenAI CUA. Our humble \model{}-4B agent is only a few points behind the specialist grounding model while gaining competence on web task completion in addition to grounding.
\section{Related work}
\label{sec:related}

We now review related work on GUI understanding, different kinds of web agents, and evaluation benchmarks for web agents.

\noindent\textbf{LLM-driven web agents.} Given the generalizable reasoning skills demonstrated by modern LLMs, many works have explored using an LLM as the core reasoning engine for a web agent, often employing prompting frameworks like ReAct~\cite{yao2023react} that interleave reasoning and action steps. This bifurcates into two broad approaches: (1) using an LLM to directly operate on a language representation of the web page derived from the DOM (such as the accessibility tree representation) to predict browser interactions~\cite{gur2023webagent,deng2024mind2web,Zhou2023WebArena,Chezelles2024BrowserGym}, or (2) exposing web capabilities via API-based tools (e.g. using search engine APIs or other website specific APIs) to a tool-use or coding agent~\cite{Song2024BeyondBA,Trivedi2024AppWorld}. While this approach works well for specific use cases where the APIs are available and known a priori, it lacks the generality of directly operating the browser as a human would, which unlocks the full range of economically useful tasks on the web. Other work has explored fine-tuning LLMs on web interaction data~\cite{lai2024autowebglm}, multi-agent orchestration frameworks that decompose web tasks across specialized agents~\cite{fourney2024magenticone}, and tree search methods for planning~\cite{koh2024tree}.

\noindent\textbf{Multimodal web agents.} In contrast, there are increasing efforts to build agents that process screenshots to produce actions. Earlier work in this direction involved creating modular systems involving planning, grounding, and verification modules~\cite{Abuelsaad2024AgentE,Zheng2024Seeact} or training dedicated visual language models for GUI interaction~\cite{hong2023cogagent,zhan2023autoui}. Recent work focuses more on a unified approach using a VLM to output actions given screenshot, instruction, and action history. This includes proprietary closed models like Gemini and OpenAI computer-use models~\cite{google2026computeruse,openai2025computeruse}, and a handful of open-weights-only models like Fara~\cite{fara7b2025}, UI-Tars family of models~\cite{Qin2025UITARS,UITARS15,wang2025uitars2}, Holo-1~\cite{Andreux2025Holo1}, and OpenCUA~\cite{Wang2025OpenCUA}. Others have explored reinforcement learning with search for web agent training~\cite{putta2024agentq}. Unfortunately, without fully open training data and transparent training and evaluation pipelines, it's impossible to advance the science of these multimodal agentic systems. Our work aims to fill this gap with our fully open training dataset \dataset, training and evaluation pipelines, and weights for our 4B and 8B \model models. It is also worth noting that, unlike prior work such as Fara, we avoid distillation from proprietary vision-based web agents. Our data pipeline largely relies on human trajectories and synthetic trajectories generated from AxTree agents that do not see screenshots.

\noindent\textbf{GUI understanding.} Separate from applications to web agents, many works have looked at the GUI parsing tasks in isolation, including GUI referring expression grounding~\cite{You2024FerretUIGM,Li2024FerretUI2M,Qian2025UGroundTU,Li2025ScreenSpotPro,Wu2025GUIActor}, answering questions about screenshots, and parsing screenshots into structured representations~\cite{Baechler2024ScreenAI,Lu2024OmniParserFP,Yu2025OmniParserVS}. While we use GUI grounding and QA data as sources of auxiliary supervision, our primary goal is to learn to follow instructions to solve tasks on the web.

\noindent\textbf{Evaluation of web agents.} Evaluating web agents is challenging. Early evaluation work focused on sandboxed web environments~\cite{Shi2017WoB,MiniWobPlusPlus,yao2022webshop,Zhou2023WebArena,Koh2024VisualWebArena}, desktop environments~\cite{xie2024osworld}, and multi-turn dialogue navigation datasets~\cite{lu2024weblinx} where the answer is known or verifiable using oracle knowledge of environment state.  Recently, several benchmarks have proposed evaluating on live websites. While some use automatic verifiers~\cite{Drouin2024WorkArena, Boisvert2024WorkArenaPP} or simple text answers that are unlikely to change over time~\cite{Yoran2024AssistantBench}, other use a VLM-as-a-judge to verify task completion correctness~\cite{He2024WebVoyager,pan2024webcanvas,Xue2025Om2w,Lyu2025DeepShop}. A VLM-judge (typically a frontier model such as GPT-4o~\cite{openai2024gpt4o}) takes the instruction, screenshots, and the final answer produced by the agent, along with a prompt specifying the success criteria, and outputs a success or failure decision, along with a rationale for that decision. 
Online-Mind2Web~\cite{Xue2025Om2w} proposes a more thorough judge that first prompts an LLM to identify key requirements for the successful completion of the task, then identifies key screenshots, and then focuses only on these key screenshots to make the final judgment.

\section{Capabilities and Limitations}
\label{sec:limitations}

We now discuss the capabilities and limitations of \model along various dimensions:

\begin{itemize}
    \item \textbf{Instruction following:} \model was trained with a wide range of instructions across different levels of specificity. The agent performs best for more specific instructions but performance may degrade as the instructions become more ambiguous or require more exploration. \model may work best when the name or URL of the target website is mentioned in the instruction. \model may also struggle to follow instructions with many constraints or search filters.
    \item \textbf{OCR and reading comprehension:} To answer questions about the current screenshot (e.g., a question pertaining to a paragraph visible in the screenshot), the agent needs to not only localize the relevant part of the webpage, but also implicitly perform OCR and reading comprehension. \model is surprisingly capable at this challenging task, but we do see instances of failures due to OCR for smaller texts or for answering complex questions requiring understanding of large passages of text. 
    \item \textbf{Latency:} \model is trained on trajectories that aim to be efficient and minimize the number of actions required to accomplish the task. However, the model is not optimized for latency (i.e., model feedforward inference time per step). Feedforward efficiency may further be optimized through techniques like quantization or pruning. Another latency bottleneck is the time it takes for the action to be executed in the browser. Browser action execution is faster for local browsers than hosted browsers like Browserbase (especially with proxy configurations), but the latter allows effortless scaling to many concurrent sessions.   
    \item \textbf{Thoughts:} An interesting capability by virtue of producing thoughts is that \model can sometimes use thoughts as memory for storing information during the trajectory and reference it for producing the final answer (since past thoughts and actions are provided as context at every step). For instance, when asked to list top news headlines, the agent would keep track of all headlines via thoughts as it scrolls through the news headlines list. However, does not demonstrate this behavior robustly. We also see failure modes where thoughts sometimes do not correlate well with actions. For instance, the thought might say the agent needs to type some text and press Enter but the action is directly to press Enter. 
    \item \textbf{Action space:} \model uses the action space that humans use when interacting with GUIs. Combining some of these actions could yield shorter trajectories. For example, for searching a query with a search box, \model currently clicks to select the input box, uses a \verb|keyboard_type| action, and then either uses \verb|keyboard_press(`Enter')| or clicks on the search button to execute the search. This sequence could be combined into a single \verb|type_at(text,x,y,press_enter=True)| action. Another potentially useful action not currently included is \verb|web_search(query)| that directly passes query arguments to a search engine via URL parameters (e.g. \cite{fara7b2025} may benefit from this action). \model struggles with predicting more infrequent actions like \texttt{scroll\_at}, \texttt{mouse\_drag\_and\_drop}, or \texttt{hover}. 
    \item \textbf{Error correction and restarts:} While \model shows some error correction behavior (e.g., using the \verb|go_back| action or scrolling to the top of the page if the agent did not find the information on scrolling down), the agent may sometimes get stuck in states where it incorrectly keeps predicting the same action (eg. repeatedly scrolling or clicking at the same location) without being able to recover or course-correct.
\end{itemize}

\section{Conclusion}
We introduced \dataset and \model, a fully open data and model suite for multimodal web agents. Operating purely from screenshots, \model agents outperform both comparable open-weight models and set-of-marks agents built on much larger proprietary models, while benefiting from test-time scaling via parallel rollouts. We'll release all checkpoints, data, code, and evaluation tools to support reproducible research and accelerate progress on open web agents.

\subsection*{Author Contributions}
\label{subsec:}
Tanmay Gupta, Piper Wolters, Zixian Ma and Peter Sushko collectively contributed to dataset construction, model training, and conducted numerous exploratory experiments for this project.

\noindent\textbf{Tanmay Gupta} conceptualized and led the project. He worked with all contributors to provide directional guidance, ideate, plan, and debug. Tanmay contributed code to all aspects of \model--data collection, training, and evaluation. He also wrote the initial versions of synthetic data generation and evaluation harnesses, created the task generation pipeline for human data annotation, and coordinated with Snorkel AI throughout the human trajectory annotation effort.  \\
\noindent\textbf{Piper Wolters} led the development and integration of the project's core infrastructure, spanning human data collection, synthetic data generation, environment setup, model training, evaluation, and the demo. \\
\noindent\textbf{Zixian Ma} led model training, grounding, and multi-agent synthetic trajectories. She generated and experimented with the grounding datasets and multi-agent synthetic trajectories. She also performed extensive experiments on training and inference setups. Together with Tanmay and Piper, she shaped the training and evaluation codebases in the beginning and contributed throughout the project. Together with the team, she trained and evaluated the final models. \\
\noindent\textbf{Peter Sushko} led the synthetic data effort and model evaluation. He optimised the axtree data generation pipeline, produced synthetic tasks, collected the single-agent axtree and node traversal datasets, and developed data filtering strategies. Together with Piper Wolters he wrote much of the model evaluation code, focusing on web environments and large-scale distributed evaluation. He ran modeling ablations on dataset mixtures and ran evaluations for many baseline models.\\
\noindent\textbf{Rock Yuren Pang} led development of the demo and scaled it for broad public deployment, implementing safety guardrails to ensure responsible use and designing the human-AI interaction. He also played a key role in improving the human annotation tool, refining its initial implementation to enhance accuracy and streamline the human-in-the-loop workflow.\\
\noindent\textbf{Diego Llanes} created the Screenshot QA dataset and contributed to the task generation and trajectory filtering pipelines.\\
\noindent\textbf{Yue Yang} explored synthetic data generation using code generation models and was involved in discussions related to data generation, training, and evaluation.\\
\noindent\textbf{Taira Anderson} managed the project, resources, timelines and helped with cross-team communication.\\ 
\noindent\textbf{Boyuan Zheng, Zhongzhen Ren, Harsh Trivedi} were involved in discussions related to training and evaluation.\\ 
\noindent\textbf{Taylor Blanton} created the initial versions of the human annotation tool.\\
\noindent\textbf{Caleb Ouellete} worked with Rock in setting up the required infrastructure for the demo.\\
\noindent\textbf{Winson Han} created beautiful figures for this technical report.\\ 
\noindent\textbf{Ali Farhadi} advised the project.\\
\noindent\textbf{Ranjay Krishna} was the PI for the project.

\subsection*{Acknowledgements}
This work would not be possible without the support of our colleagues at Ai2:
\begin{itemize}
    \item We thank the PRIOR team for helpful research discussions and sharing of findings across related projects.
    \item We thank Malachi Hamada as well as Snorkel AI for their contributions to the human annotation efforts.
    \item We thank David Albright, Crystal Nam, Will Smith, David Everhart, Kyle Wiggers and Cailin Brashear for their work on public release of MolmoWeb.
\end{itemize}

\clearpage

\bibliography{main}

\clearpage

\appendix
\appendix

\section{Overview}

In this appendix, we provide the following:
\begin{itemize}
    \item Details for human (\cref{sec:supp_human})and synthetic (\cref{sec:supp_synthetic}) trajectory generation pipelines, including details and prompts for how the tasks are sampled. 
    \item Additional dataset statistics like distribution of web categories, websites, and actions covered in \dataset. (\cref{sec:supp_data_analysis})
\end{itemize}

\section{Human Trajectory Annotation}
\label{sec:supp_human}

\subsection{Annotation Tool}
\begin{figure}[tb]
  \centering
    \includegraphics[width=\linewidth]{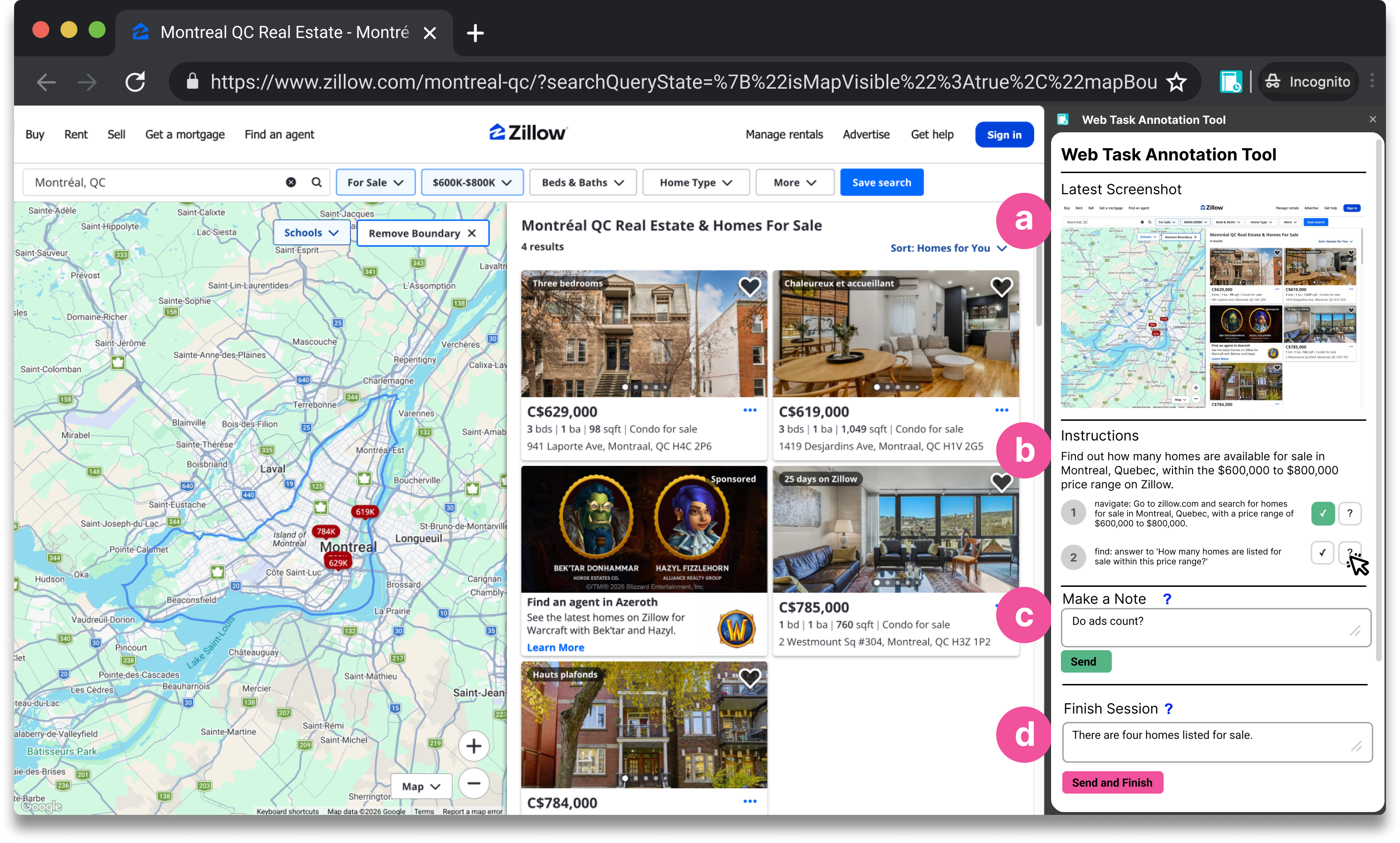}
  \caption{Screenshot of the web browsing trajectory collection tool. The left half of the screen shows the webpage, the panel on the right is our chrome extension with annotations showing: \textbf{(a)} last captured screenshot; \textbf{(b)} instruction given to the annotator with a step-by-step breakdown; \textbf{(c)} input box for leaving a note when marking a step could not be completed; \textbf{(d)} final answer.}
  \label{fig:extension}
\end{figure}

Collecting high-accuracy human trajectories from the annotation tool (\Cref{fig:extension}) is challenging due to various quirks of the Chrome extension, particularly around capturing screenshots along with DOM events. These timestamped events and screenshots were post-processed into a clean sequence of screenshots and user actions taken on those screenshots. Annotation workers were given detailed guidelines to ensure all important actions are captured. Particularly, the workers were asked to wait long enough after each action for the webpage to complete loading and for the annotation tool to trigger a screenshot automatically, or to manually take a screenshot in case an automatic screenshot is not triggered. Workers were also discouraged from acting too quickly to avoid missing events and screenshots via interventions like displaying a warning message and through annotation training. We partnered with Snorkel AI to scale-up the human annotations. Specifically, we provided tasks and our annotation tool to Snorkel AI who then managed the annotation workers, verified the annotations for correctness, and ensured quality control. 

\subsection{Task Sampling}
To ensure a diversity of tasks for workers to annotate, we generate tasks using multiple strategies as described below:

\subsubsection{Manually written task templates.} To generate a core set of common use cases, authors wrote task templates. These include tasks related to shopping, news, real estate, travel (including flights, car rentals, hotels, etc.), maps, food \& recipes, job \& salary search, health \& wellness, and cars. Each task template consists of a sequence of atomic skills (using the taxonomy defined in Tab.1 in the main paper) with placeholders that are populated with samples drawn from predefined allowed constants (e.g., destinations for flight search) or sometimes generated on the fly (e.g., filters to apply while shopping for a product). Here's an example for a shopping search and filter task:

\begin{verbatim}
go to: walgreens.com
search: coffee
apply filters: brand=Lavazza, availability=Pickup
find and open: most relevant product
\end{verbatim}

The annotation tool requires the workers to check each of these steps as they are completed. If a step can not be completed, the tool allows the worker to mark the step as incomplete along with a reason why the step could not be completed (e.g., a filter does not exist).

For training purposes, in addition to these step-by-step instructions, we use an LLM to rewrite these instructions with different levels of specificity. The most specific low-level instruction is just a paraphrase of atomic steps into colloquial English, mid-level instruction is less verbose and could leave out details that may be obvious from context, and the highest-level instruction just states the intent without necessarily prescribing a way to achieve the intent. Here is an example of the different levels of specificity:

\begin{itemize}
\item \textbf{Low-level}: Go to walgreens.com, search for coffee, apply filters for Lavazza brand and available to pickup, and open the most relevant product.
\item \textbf{Mid-level}: On walgreens, search for Lavazza coffee available for pickup, and open the closest matching product.
\item \textbf{High-level}: Find Lavazza coffee for pickup at walgreens.
\end{itemize}

During training, we randomly sample one of the 4 available instructions (steps-by-step and 3 levels) for each trajectory using a slightly higher sampling probability for the high-level instruction.

\subsubsection{LLM sampled tasks with steps.} To increase diversity, we also sample instructions using an LLM. We use a persona sampled from PersonaHub~\cite{Chan2024PersonaHub} to amplify task diversity along with the following prompt:

\begin{minipage}{\linewidth}
\begin{lstlisting}
You are {persona}. Your goal is to generate interesting web tasks that can be performed on the given website. The task must be specified first as sequence of allowed steps and then as a natural language instruction to a web agent.

# Allowed steps
Here are the allowed steps with descriptions -
- go_to: navigate to the website url using url bar. Takes the website url as argument
- search: searching using a simple search box like on google.com or amazon.com. Takes the search string as argument. Only use this if you know there is a search box on the page.
- find: scroll to locate relevant information in the current page. Takes a description of what to find as argument
- find_and_open: find and open the relevant page. Takes a description of what to find and open as an argument. Could be specific like "shopping cart" or general like "most relevant search result"
- find_and_click: find a relevant element on the page and click on it. Takes a description of what to find and click as an argument.
- fill_form: fill a form like on airbnb.com or southwest.com. Takes form details as argument. Useful when you need to take some other action like applying filters before submitting the form
- fill_form_and_submit: fill the form and submit it too. Takes form details as argument
- apply_filters: apply filters like product related filters on amazon but don't execute the search yet. Takes the filter keys and values are arguments
- apply_filters_and_search: apply filters and run the search. Takes filter keys and and values as arguments
- add_to_cart: when on the product page, add it to cart (for products) or to booking/reservation (for hotels, flights etc). Takes the amount to add to cart as argument

# What kinds of task to generate?
1. Generate tasks that require navigating to a target page and optionally answering a question based on the target page. When requiring question answering, the last step should be find: answer to "{{question}}"
2. Typically the tasks should have 3-10 steps
3. Task should neither be too easy nor too complex. Generally doable in 5-10min
4. Some websites may provide advanced search functionality - make use of these in your tasks whenever possible
5. When providing dates for booking or reservation tasks provide dates in 2026. For tasks that require searching existing data provide dates before 2025 or provide relative dates like "last 30 days", "next 2 days", "a week from now" etc
6. Do not generate tasks that require uploading files, or opening pdfs
7. Do not generate tasks requiring logins, or user's personal information like name, address, credit card details etc
8. Generate website relevant tasks, specially search keywords, filters, and form details

WEBSITE: {url}
\end{lstlisting}
\end{minipage}

Similar to the manually written steps, we generate instructions at 3 levels of specificity for training from these LLM-sampled step-by-step tasks.

\subsubsection{LLM sampled tasks with navigation and QA.} While well structured and easy to follow, the tasks generated by the above prompt can be too rigid and may sometimes be infeasible (e.g., strict search filter constraints result in no matches on the website). 
To generate more flexible tasks, for a portion of the tasks, we relax the constraint of generating sequence of atomic steps, and ask the LLM to generate a flexible request to navigate to a target webpage, followed by an optional question to be answered once on the target page. Here's an example of such a task:

\begin{verbatim}
navigate: Find Lavazza coffee for pickup on Walgreens
question: How much does it cost?
\end{verbatim}

The prompt used for sampling these navigation and QA tasks was as follows:

\begin{minipage}{\linewidth}
\begin{lstlisting}
You are {persona}. Your goal is to generate interesting web tasks that can be performed on the given website. The task must be specified first as sequence of allowed steps and then as a natural language instruction to a web agent.

# Allowed steps
Here are the allowed steps with descriptions -
- navigate: a self contained instruction to navigate to a target webpage. This could involve using a search engine, navigation bar, directly navigating to a url, as well as performing simple or advanced search on websites (like Github, Arxiv), and/or opening a particular search result, shopping cart, etc. When targeting a specific website, include the name of the website (if popular) or the URL of the website.
- question: a question about the target page

# What kinds of task to generate?
1. Generate tasks that require navigating to a target page and optionally answering a question based on the target page.
2. There are only two kinds of tasks to generate: navigation only tasks that only contain a single navigation step; and 2 step task that contain a navigation step followed by a question.
3. Task should neither be too easy nor too complex. Generally doable in 5-10min
4. Some websites like Github, Arxiv, Google Scholar etc may provide advanced search functionality - make use of these in your tasks whenever possible
5. When providing dates for booking or reservation tasks provide dates after March 2026. For tasks that require searching existing data provide dates before 2025 or provide relative dates like "last 30 days", "next 2 days", "a week from now" etc
6. Do not generate tasks that require uploading files, or opening pdfs
7. Do not generate tasks requiring logins, or user's personal information like name, address, credit card details etc
8. Generate website relevant tasks, specially search keywords, filters, and form details

WEBSITE: {url}
\end{lstlisting}
\end{minipage}

\subsubsection{LLM sampled benchmark-like tasks.} All the above generation methods are agnostic to the target evaluation benchmarks and generally aim to cover diverse websites and tasks. However, the above approaches contain biases of authors (either directly in manually written templates or indirectly through prompts) that are likely to be different from the biases of the downstream benchmarks. Therefore, to close the gap in the distribution of tasks in the benchmarks, we generate a set of tasks using tasks from two of the four target benchmarks as in-context examples. Specifically, given a website and all tasks for that website from WebVoyager and Online-Mind2Web, we ask an LLM to generate tasks that are similar to the in-context examples, but not simply paraphrases.

We use GPT-4o for the majority of the LLM sampled tasks, but for some tasks we cycle between GPT-4o, GPT-4.1, GPT-5-mini, and GPT-5 to avoid biases of any single LLM and to increase task diversity.

\section{Synthetic Trajectories Generation}
\label{sec:supp_synthetic}
\subsection{Task Sampling}

Sampling tasks for synthetic trajectories use the following strategies:

\subsubsection{Manually written task templates.} Similar to manually written task templates for human annotation, a portion of tasks for generating trajectories are manually authored with detailed task templates. The main difference from the manual templates for human annotation is that these are natural language templates without a very well-defined atomic skill taxonomy. These manually written templates focused mainly on the websites in the WebVoyager benchmark.

\begin{figure}[t]
  \centering
    \includegraphics[width=\linewidth]{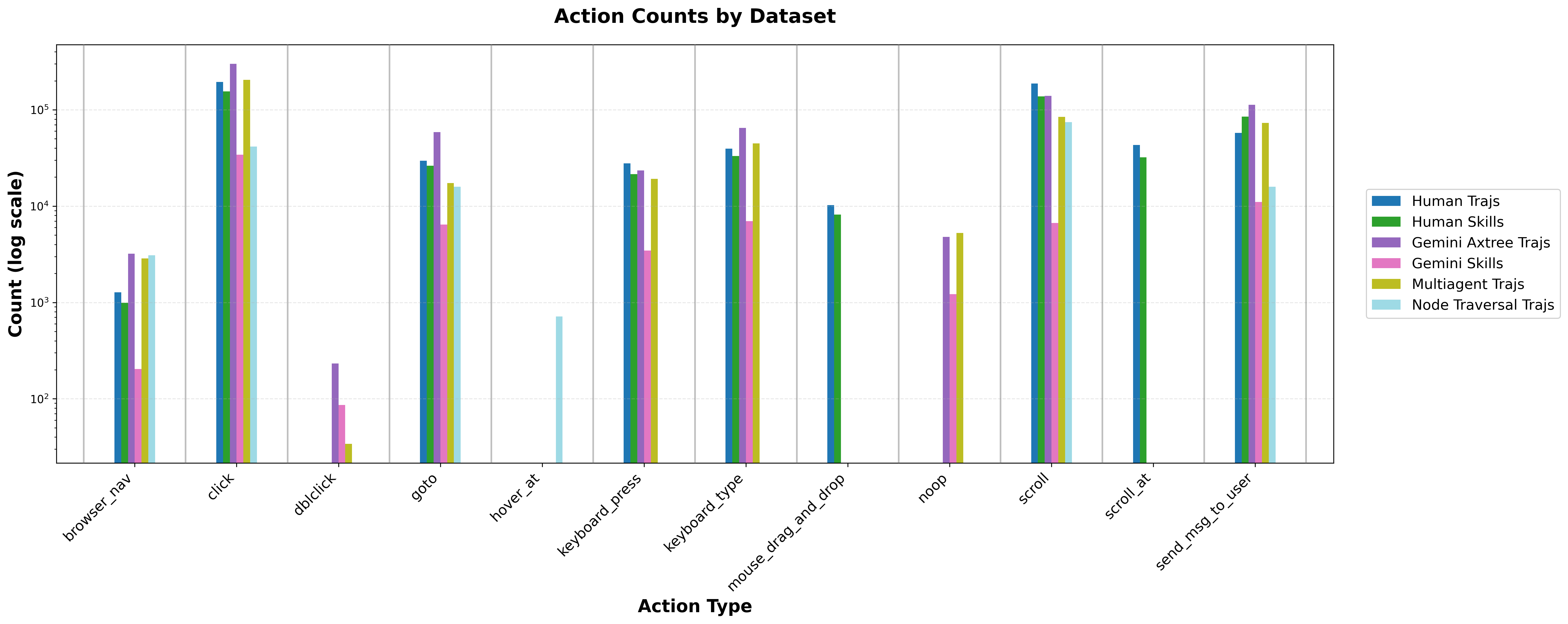}
  \caption{Action distribution breakdown across the dataset.}
  \label{fig:action_dist}
\end{figure}

\begin{figure}[thb]
  \centering
    \includegraphics[width=0.9\linewidth]{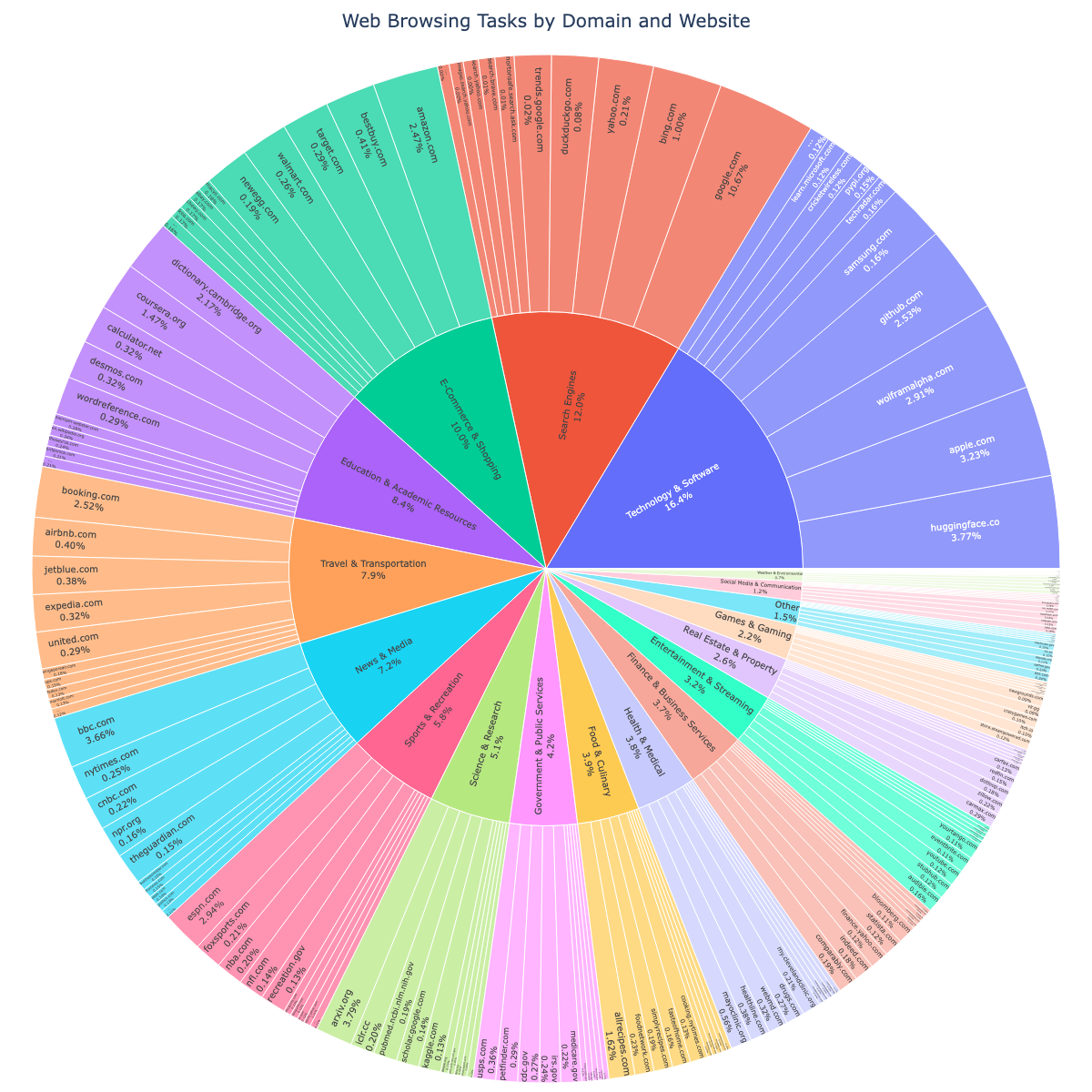}
  \caption{Sunburst chart of domains and websites in the dataset.}
  \label{fig:sunburst}
\end{figure}

\subsubsection{LLM sampled benchmark-like tasks.} In a manner similar to the LLM sampled benchmark-like tasks for human annotations, we generated tasks that match the distribution of WebVoyager. For Online-Mind2Web, for a given task, instead of just generating tasks for the original website, with some probability, we sample a website from the same category (e.g., if the original website is a shopping website, we sample another shopping website from a pre-categorized list of websites), and ask an LLM to generate a similar task for that website.

\subsubsection{Taxonomy-based tasks generation}
To enable generating diverse tasks and supplementing existing tasks, we define a comprehensive task taxonomy across four axes: (1) intent – what the user wants to do: info-seeking (looking up information), transactional (purchasing/booking), tool-use (using web tools to compute or create), messaging (sending messages or posting content), and navigation (reaching a specific page); (2) domain – the website category, spanning 13 areas: travel, ecommerce, productivity-apps, communication-apps, devtools, social, finance, education, media, reference, local, gov, and health; (3) difficulty – task complexity by constraint count: D0 (0–1 constraints), D1 (2–3), D2 (4–5), D3 (6+). Higher levels require satisfying more simultaneous requirements; (4) ambiguity – goal clarity and solution space: A0 (single correct answer), A1 (optimize under constraints, multiple valid solutions), A2 (open-ended comparison or recommendation), A3 (vague/underspecified goals needing clarification).

 We then develop a pipeline to generate synthetic web tasks by systematically iterating over this taxonomy. For each cell in the Cartesian product of these axes, it prompts an LLM (i.e., GPT-4o) to produce a specified number of diverse, realistic web browsing tasks. Each generated task includes a target website, a task type label, a sequence of concrete browser action steps, and three verbosity levels of natural language instructions (high-level, mid-level, low-level). The prompt constrains the LLM with detailed descriptions of each taxonomy axis value and optionally restricts website choices using a list of curated websites. Tasks are tagged with their taxonomy metadata during the generation process. 

\section{Additional Dataset Statistics}
\label{sec:supp_data_analysis}

Across our synthetic and human datasets, a wide variety of web browsing domains and websites are covered.  \Cref{fig:sunburst} presents a sunburst chart illustrating the hierarchical breakdown of broad web domains and websites represented in our training data, with just the top 9 websites for each category listed for readability. Our datasets are constructed to closely approximate the action space and input distribution characteristic of human web browsing behavior. In \Cref{fig:action_dist} we show the distribution of actions across our various training datasets.

\end{document}